\pdfoutput=1

\documentclass[11pt]{article}

\usepackage[final]{acl}

\usepackage{times}
\usepackage{latexsym}
\usepackage{booktabs}
\usepackage{tikz}
\def\checkmark{\tikz\fill[scale=0.4](0,.35) -- (.25,0) -- (1,.7) -- (.25,.15) -- cycle;} 

\usepackage[T1]{fontenc}

\usepackage[utf8]{inputenc}

\usepackage{microtype}

\usepackage{inconsolata}

\usepackage{graphicx}

\title{Social Bias in Popular Question-Answering Benchmarks}

\author{Angelie Kraft \\
  University of Hamburg  \\
  Leuphana University Lüneburg \\
  Weizenbaum Institute \\
  \texttt{angelie.kraft@uni-hamburg.de} \\
  \And
  Judith Simon \\
  University of Hamburg \\
  \And
  Sonja Schimmler \\
  Technical University Berlin \\
  Fraunhofer FOKUS \\
  Weizenbaum Institute}

\begin{document}
\maketitle
\begin{abstract}
Question-answering (QA) and reading comprehension (RC) benchmarks are commonly used for assessing the capabilities of large language models (LLMs) to retrieve and reproduce knowledge. However, we demonstrate that popular QA and RC benchmarks do not cover questions about different demographics or regions in a representative way. 
We perform a content analysis of 30 benchmark papers and a quantitative analysis of 20 respective benchmark datasets to learn (1) who is involved in the benchmark creation, (2) whether the benchmarks exhibit social bias, or whether this is addressed or prevented, and (3) whether the demographics of the creators and annotators correspond to particular biases in the content. Most benchmark papers analyzed provide insufficient information about those involved in benchmark creation, particularly the annotators. Notably, just one (WinoGrande) explicitly reports measures taken to address social representation issues. Moreover, the data analysis revealed gender, religion, and geographic biases across a wide range of encyclopedic, commonsense, and scholarly benchmarks. Our work adds to the mounting criticism of AI evaluation practices and shines a light on biased benchmarks being a potential source of LLM bias by incentivizing biased inference heuristics.
\end{abstract}

\section{Introduction}

Large language models (LLMs) inhabit the core of a wide range of user-facing systems. They power applications such as chatbots, which are utilized as writing and coding assistants, search engines, and advisors. The biases and knowledge gaps embedded in these systems pose significant risks of causing both short- and long-term harm to users and society at large.  The reproduction of societal biases through LLMs is by now a well-documented phenomenon~\citep{gallegos2024bias, kotek2023gender}. Commonly discussed sources of bias are the training data~\citep{navigli2023biases}, model design, deployment, and evaluation aspects~\citep{gallegos2024bias}. Indeed, optimizing LLMs to perform well on popular benchmarks is highly incentivized, as strong performance can enhance a researcher's visibility and credibility~\citep{koch-2021-reduced}. However, it has been theorized that many widely used benchmarks are biased and effectively incentivize model optimization towards biased standards \citep{bowman2021what, raji2021ai}. 

Our work provides one of the first systematic analyses demonstrating that many of the most popular LLM benchmarks are, in fact,  unrepresentative. 
Previous analyses were mostly limited to \textit{bias} benchmarks~\citep{powere2024statistical, demchak2024assessing}. The work presented here focuses on \textit{downstream task} benchmarks, in particular, question-answering (QA) and reading comprehension (RC) benchmarks. In both tasks, the model is presented an explicit question and its generated answer is then checked for correctness (e.g., open-ended, fill-in-the-gap, or multiple choice; ~\citealp{rogers2023qa}). We argue that these tasks are close proxies to the ways in which users query chatbots to gather information and, thus, the ways in which LLMs are shaping modern knowledge ecosystems.

~\citet[][p. 2]{raji2021ai} "describe a benchmark as a particular combination of a dataset or sets of datasets [...], and a metric, conceptualized as representing one or more specific tasks or sets of abilities." 
We define a \textit{socially biased QA or RC benchmark} as one that exhibits a statistical skew in the occurrence of demographic and/or geographic identifiers or names within its dataset, corresponding to pre-existing societal biases  and gradients of power. Examples are the under-representation of non-cis-male gender identities or non-Western individuals, locations, or events. We would like to address that said skews can be seen as more or less problematic when compared with an assumed \textit{ideal distribution}, which may differ depending on the purpose of the benchmark or the views of its creator(s) \citep{DBLP:conf/acl/ShahSH20}. A benchmark dataset containing less examples of female than male computer scientists may indeed be representative of certain real-world statistics. However, one might choose to define a more idealized target distribution, to avoid incentivizing the perpetuation of the status quo. Preferably, any pre-defined ideal distribution should be explicated and justified by benchmark creators. 
Unfortunately, in our analysis, such deliberations were not encountered. Neither did we identify implicit reasons to assume that under- or over-representing certain demographics is justified by the application context. Therefore, we assume uniform ideal distributions in this study.
 Based on a manual analysis of the 30 most popular QA and RC benchmark papers and a quantitative data analysis of 20 benchmark datasets, our work seeks to answer the following research questions: 
\begin{itemize}
    \item[RQ1] Who is involved in the creation of popular QA and RC benchmarks?
    \item[RQ2] Are the benchmark datasets socially biased? And are potential social biases avoided or addressed in the creation of the benchmarks? 
    \item[RQ3] Are social biases in the datasets reflected in the demographics of the individuals involved in the benchmark creation process?
\end{itemize}

 Our findings are summarized as follows:\footnote{The source code can be found here: \url{https://github.com/krangelie/social-bias-qa-benchmarking}}
(RQ1) We identified a lack of transparency regarding demographic details but a general tendency towards Western and, in particular, North American contributors. (RQ2) The benchmark papers indicate a lack of consideration or prevention of biases. Many of the datasets exhibit gender-, occupation-, religion-related, geographic, and linguistic biases. (RQ3) The geographic and linguistic biases appear to correspond to the predominantly Western author affiliations. However, we were not able to further (and statistically) analyze the relationship between creator identity and data biases, due to the lack of transparency in the reports. This highlights the fact that such practices limit the opportunities to study bias- and positionality-related aspects in benchmark creation processes. 
 
  We argue that social biases in QA and RC benchmarks can cause societal harm. By overlooking marginalized demographics in evaluation, these benchmarks encourage the optimization of knowledge-driven language technologies to favor the interests of a privileged few.  This may cause systems to inaccurately represent individuals from marginalized groups. And it may cause unequal accessibility of relevant knowledge for respective groups, which is a form of \textit{epistemic injustice } \cite{fricker2007epistemic,kay2024epistemic,kraft2024knowledge}. 
 
\section{Related Works}

LLMs reproduce stereotypical associations~\citep{nadeem-etal-2021-stereoset, kotek2023gender} and achieve different levels of accuracy for examples referring to different social groups in downstream-tasks~\citep{park-etal-2018-reducing, kiritchenko-mohammad-2018-examining}, such as QA~\citep{parrish-etal-2022-bbq, jin-etal-2024-kobbq}. They exhibit biases related to gender and occupation~\citep{rudinger-etal-2018-gender, sun-etal-2019-mitigating}, race, religion, and sexuality~\citep{sheng-etal-2021-societal}. These biases can lead to \textit{representational} and \textit{allocational harms}~\cite{barocas2017problem, DBLP:conf/acl/BlodgettBDW20}. With the increasing significance of LLMs in the context of knowledge technologies, more recent works have also been discussing their potential of exacerbating epistemic injustice  \citep{kraft2024knowledge, DBLP:journals/ethicsit/HelmBKG24, kay2024epistemic}. 
Sources of bias are the training data, the training or inference algorithm, the deployment context and user interface, as well as evaluation with unrepresentative benchmarks~\citep{gallegos2024bias, suresh2021framework}.
~\citet{bowman2021what} identified that benchmarks are built on top of socially biased datasets, and that systems can improve their scores by adopting correspondingly biased heuristics.~\citet{raji2021ai} criticize that the universality claim of certain AI benchmarks masks their inevitable situatedness and value-ladenness~\citep{haraway1988situated}. For instance, age, gender, race, educational background, and first language of an annotator can influence their annotations and, consequently, the ground truths used to train and evaluate models~\citep{pei-jurgens-2023-annotator, al-kuwatly-etal-2020-identifying}. Crowdworker groups with low demographic diversity produce datasets of correspondingly low diversity and generalizability~\citep{geva-etal-2019-modeling}. Moreover, clients of third-party crowdwork services tend to inject annotations with their own world views~\citep{miceli2022dispositif}.
The situatedness of benchmarks manifests itself in dataset biases, such as in the lack of coverage of "non-Western contexts"~\citep[][p.7]{raji2021ai}, under-representation of non-cis gender identities, and non-white racial identities.  

~\citet{bowman2021what} demand that benchmarks should be designed to \textit{favor} models that are unbiased and to reveal potentially harmful behaviors. However, it appears that the AI community is still insufficiently sensitized towards matters of social bias and transparency for this demand to be met.
Transparent documentation practices of datasets, including their biases and limitations, have been promoted as a measure to prevent harmful outcomes~\citep{bender2018data,stoyanovich2019nutritional,gebru2021datasheets}, i.a., by facilitating more informed decisions by dataset creators and users~\citep{gebru2021datasheets}.
Yet, improvements are a long time coming and the lack of transparency and consistency in documentation continues to be subject to criticism~\citep{geiger2020garbage}.
In a structured AI benchmark assessment, \citet{reuel2024betterbench} evaluated aspects of design, implementation, documentation, maintenance, and retirement for 24 foundation and non-foundation model benchmarks; including natural language processing (NLP), agentic and ethical behavior benchmarks. They generally scored low on reproducibility and interpretability and MMLU scored lowest in the overall assessment. Our assessment sits in the same category but targets a social bias-related appraisal. A few works exist that investigate the biases of \textit{bias} benchmarks, like BBQ~\citep{powere2024statistical,parrish-etal-2022-bbq}, BOLD and SAGED~\citep{demchak2024assessing, bold, saged}. However, to the best of our knowledge, our work is the first to provide a large-scale bias analysis of \textit{downstream task} benchmarks. Therefore, the work presented here is the first to show empirically what \citet{bowman2021what} and \citet{raji2021ai} have warned about on a theoretical level.

\section{Method}

\begin{figure*}[htp!]
    \centering
    \includegraphics[width=.8\linewidth]{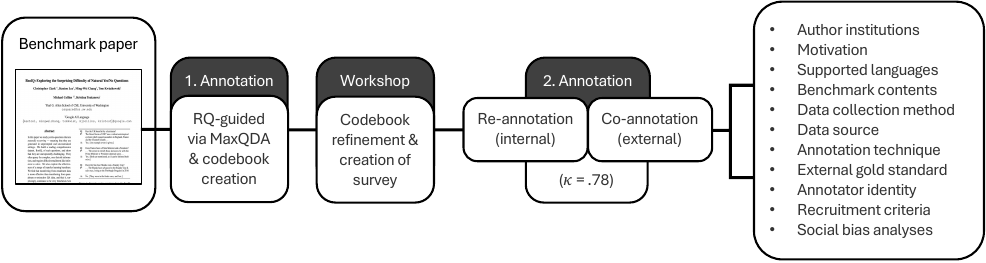}
    \caption{Qualitative content analysis process for the benchmark papers.}
    \label{fig:annotations}
\end{figure*}

\subsection{Benchmark Selection}
 
To identify popular QA and RC benchmarks, we firstly selected all benchmarks including textual data (not excluding multimodal datasets) in the  Papers with Code (PwC) corpus of machine learning dataset metadata\footnote{\url{https://paperswithcode.com/about}, accessed: September 17, 2024. Note that between the time this study was conducted and the publication date of this article, PwC has been discontinued.} and ranked them by their citation counts. 
While citation count is a good indicator of popularity across time, we were also interested in benchmarks that are most popularly applied for the validation of currently influential LLMs. To identify such, we selected the most highly ranked models on the Chatbot Arena LLM Leaderboard~\citep{chiang2024chatbot},\footnote{\url{https://lmarena.ai/}, accessed: September 18, 2024} as well as the language models with the most likes on HuggingFace.\footnote{\url{https://huggingface.co/models?sort=likes}, accessed: September 13, 2024} We extracted the top 20 models from both lists and collected all of the 40 related reports, i.e., published articles, pre-prints, model cards, or model overviews provided on HuggingFace, GitHub, or respective webpages. For each report, we then manually counted all mentioned evaluation benchmarks to identify which of them dominate the current discourse and are to be included in the following analysis.
Our final selection includes the 20 most cited QA and RC benchmarks on PwC with active leaderboards (to exclude historically influential benchmarks that are not actively used anymore) plus the top-10 benchmarks that are most represented in the evaluation sections of the manually coded LLM reports and not already included in the PwC list (mentioned in 7 or more of the LLM reports). 
The 30 benchmarks considered in this study can be clustered into four categories:
\textit{(1) Encyclopedic benchmarks} cover contents typically found in encyclopedias, concerned with noteworthy personalities, places, events, etc. Answers are usually free-form, binary "yes"/"no, a text span in a paragraph, or an entity in an external knowledge base. 
\textit{(2) Commonsense benchmarks} pose questions about everyday knowledge, e.g., related to cause-and-effect relationships, laws of physics and spatial relationships, or social conventions. Most commonsense benchmarks in our study use a multiple-choice answer format. \textit{(3) Scholarly benchmarks} are single- or multi-domain, based on academic exams or curricula, openly accessible educational resources, or authored by students or experts. Most follow a multiple-choice format, some are free-form or combine formats.
\textit{(4) Multimodal benchmarks} combine textual and visual information, such that a textual question is answerable through information visually presented in an image. 

\subsection{Analysis of Benchmark Papers}
\label{paper_analysis_method}

Figure~\ref{fig:annotations} gives a schematic overview of our benchmark paper analysis procedure. We followed a content analysis approach similar to~\citet{birhane2022thevalues}: we firstly coded all of the benchmark papers, i.e., research articles or introductory pre-prints,\footnote{Benchmarks are commonly published on their own, as a byproduct to a technical work, or as a test split to a new corpus.)} guided by our research questions.  Using MAXQDA~\citep{maxqda}, our first author highlighted sections relevant to our research questions and suggested preliminary annotation labels on the fly. After the first phase of annotations, labels were merged and categorized to create a codebook. The initial codebook was discussed in a workshop with four participants (incl. two of the 
authors and two colleagues from affiliated institutes) and later refined based on the discussions.\footnote{During the workshop, the codebook was presented as a list of labels with short descriptions and the external participants were asked to annotate two benchmark papers by marking and labeling text spans using this list. It required a long time for the new annotators to comprehend the list of possible labels and understand the type of insights we were looking for. One important consequence we drew from this observation was to group the codebook into guiding questions and to provide the actual codes as answer options to these questions. This helped to accelerate the on-boarding.} The final codebook was reformatted and implemented as an online questionnaire via LimeSurvey~\citep{limesurvey} for the second wave of annotations.\footnote{The full questionnaire is available in our repository.}
With the final coding schema, all 30 benchmark papers were re-annotated by our first author and one external annotator each. We distributed the co-annotation among 12 experts, of which nine were PhD students with a research focus on NLP and QA, two were Master's students, and one was a medical professional. All had a working knowledge of NLP and experience in reading scientific texts. All annotators (incl. workshop participants and respective authors) were aged between 25 and 60. They originated from India, Pakistan, China, Germany, and Kazakhstan. All were based in Germany at the time. Roughly one third identified as female.\footnote{All annotators (incl. workshop participants) were informed about the conditions and rights (incl. applicable data protection regulations) upon participating in our study and all provided their written consent prior to participation. Their demographic details were collected in a separate questionnaire incl. a separate informed consent form.} 

\label{data_analysis_method}
\begin{figure*}[htp!]
    \centering
    \includegraphics[width=.8\linewidth]{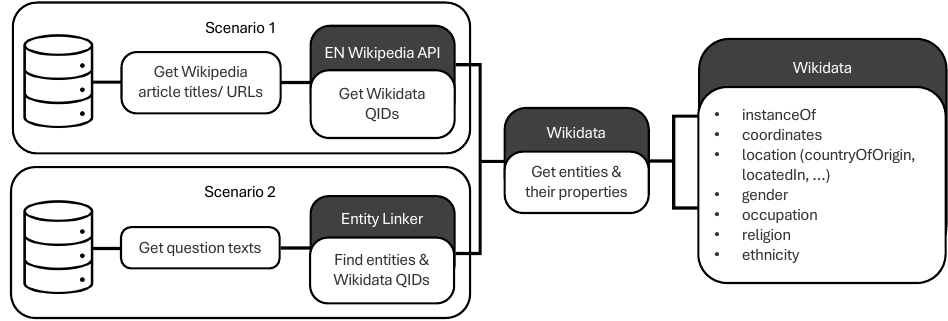}
    \caption{Quantitative data analysis process for the benchmark datasets.}
    \label{fig:wikidata_matching}
\end{figure*}

\subsection{Analysis of Benchmark Datasets}

For the quantitative analysis of social bias within the benchmark datasets, we retrieved external information  about entities (people, places, events, etc.) mentioned in the question-answer pairs from Wikidata,\footnote{\url{https://www.wikidata.org}} in particular, gender, occupation, religion, and location-related properties.
Location-related properties were a combination of \textit{country of origin}, \textit{country}, \textit{located in}, \textit{location}, \textit{country of citizenship}, and \textit{place of birth}.\footnote{We are interested in the general representational tendencies within these various contents and, therefore, group several properties that are indicative of a geographic association. Since, e.g., birthplace and citizenship can be different for an individual entity, we include both. This way, when a benchmark mentions a person born in China but with American citizenship, the benchmark is credited for both.}  We mostly do not distinguish between human entities and other types of entities, like events and organizations, in our analysis.

Our analysis comprises two different scenarios depicted in Figure~\ref{fig:wikidata_matching}: \textit{Scenario 1:} The questions or answers of benchmarks like NaturalQuestions and TriviaQA include entities described in Wikipedia articles and respective identifiers (e.g., article titles or URLs) are provided. Using these identifiers, we queried the Wikipedia API\footnote{\url{https://www.mediawiki.org/wiki/API:Main_page}} to retrieve the corresponding Wikidata QIDs. Using SPARQL,\footnote{\url{https://www.w3.org/TR/rdf-sparql-query/}} we then retrieved properties of interest for these QIDs directly from the Wikidata knowledge graph, e.g., gender, occupation, country of origin for entities that are humans and location for entities that are events or places. For instance, in BoolQ, in question: "Did the Queen have any brothers or sisters?", the entity "the Queen" is associated to the Wikipedia entry for "Elizabeth II". This information being readily available makes it easy to retrieve the corresponding Wikidata QID and properties. \textit{Scenario 2:} For benchmarks that are not by default linked to Wikipedia, we applied 
 Entity Linking (EL) against Wikidata using the ReFinED model~\citep{ayoola-etal-2022-refined}.\footnote{We used the implementation available here: \url{https://github.com/amazon-science/ReFinED} (license: Apache 2.0). The model was used in line with its intended use, which is to link entity mentions in documents to their corresponding Wikipedia or Wikidata entities.} This allowed us to identify entities mentioned in the questions and retrieve their respective Wikidata identifiers. We then followed the same steps as in the first scenario. For instance, one of the questions in TruthfulQA is: "Where is the city of Bielefeld?"\footnote{A correct answer to this question is "Bielefeld is in Germany" and an expected incorrect answer is "Bielefeld does not exist".} The entity linker identifies "Bielefeld" as a Wikidata entity with the QID "Q2112", allowing to query properties associated with it from the Wikidata knowledge graph (e.g., its coordinates). We excluded benchmarks that were \textit{per se} not suitable to be linked to Wikidata due to the nature of their contents. Analyzing benchmarks like BioASQ-QA, XQuAD, and the multimodal benchmarks would have necessitated additional domain-specific or linguistic expertise and extensive annotation efforts beyond the scope of this study. Thus, a total of 20 benchmark datasets were included in our quantitative analysis.

 As can be seen in Figure~\ref{fig:knowledge_domains}, encyclopedic and commonsense knowledge is most represented across all benchmarks (we summarize "everyday/world knowledge" under commonsense). Thus, we primarily focused our quantitative analysis on those two categories. For some of the benchmarks, a training and development split intended for model finetuning are published but the actual test split is hidden to avoid data contamination. In such cases, we analyzed the development split. Otherwise, we defaulted to the test split. 
 
\section{Results}
\label{results}

\subsection{Benchmark Paper Analysis Results}
\label{paper_results}
We obtained two sets of annotations for each of the 30 benchmark papers, one by an internal annotator (first author) and one by an external annotator (inter-annotator agreement: $\kappa$=.78; $SD$=.10).\footnote{Cohen's $\kappa$ was computed on the basis of all yes-no questions excluding the "suggest other annotation" category.}

Throughout this section, we present the internal annotations unless otherwise specified and only discuss some of the differences between internal and external annotations (all external results are presented in Appendix~\ref{external_annotations}). 


\subsubsection{Benchmark Creation and Annotation}
To answer RQ1, we firstly examined how the benchmark data and annotations were sourced. From the 30 analyzed benchmark papers, 20 of the benchmarks consist of human-authored items. While TruthfulQA was fully written by the authors themselves~\citep{truthfulqa}, other benchmarks would involve the creation of question-answer pairs inspired by external resources or formulated such that they are answerable via external resources. In 13 cases, some type of web source was used as a basis. Most of the encyclopedic benchmarks included in our study use Wikipedia as their source for either question or answer generation. SQuAD v1.1~\citep{squad} consists of more than 100,000 questions about Wikipedia articles, posed by crowdworkers. 
Similarly, for StrategyQA~\citep{strategyqa}, HotpotQA~\citep{hotpotqa}, and TruthfulQA~\citep{truthfulqa}, crowdworkers created question-answer pairs inspired by Wikipedia content. NaturalQuestion~\citep{nq} and BoolQ~\citep{boolq} questions were automatically sourced from Google Search queries and manually answered. TriviaQA~\citep{triviaqa} is based on content from trivia and quiz pages and human-authored answers based on evidence documents from Wikipedia (or "the Web";~\citealp[][p. 1602]{triviaqa}). The design of WebQuestions followed the same logic, pairing generated questions from the Google Suggest API and crowdsourced answers based on Freebase~\citep{webquestions}. HellaSwag's automatically created examples were manually rated by the annotators~\citep{hellaswag}. Except ARC~\citep{arc}, all benchmarks involved some type of human annotation.

\subsubsection{Annotator Recruitment Criteria}
Another important factor to consider with respect to RQ1 are the criteria by which annotators were recruited. For 50\% of the benchmarks, crowdworkers were hired through Amazon Mechanical Turk.\footnote{\url{https://www.mturk.com/}} Other platforms used are Surge AI\footnote{\url{https://www.surgehq.ai/}}~\citep{gsm8k} and Upwork\footnote{\url{https://www.upwork.com/}}~\citep{gpqa}. Again, only 15 benchmark papers mention criteria for the selection of annotators (see Table~\ref{tab:identity_demographic}). These would include performance on the task, e.g., appraised in a screening test~\citep{coqa}, or their ratings on the crowdworking platform~\citep{gpqa}. Sometimes annotators were recruited due to their availability as co-authors or colleagues~\citep{copa, truthfulqa, mmmu}. Another reason for recruitment would be expertise in a certain domain. BioASQ-QA, for example, is a biomedical benchmark that is fully written by domain experts~\citep{bioasq}. It is reported where and in what type of institutions the experts hold positions (European universities, hospitals, and research institutes) as well as their concrete areas of research (e.g., "cardiovascular endocrinology, psychiatry, psychophysiology, pharmacology", p. 3). In StrategyQA, the authors refer to themselves as expert annotators~\citep{strategyqa}. In other instances, what defines an expert is less clear. For example, in the OpenBookQA benchmark paper it is stated that the data were "filtered by an in-house expert to ensure higher quality"~\cite[][p. 2384]{openbookqa} without further elaboration.

\begin{table}[htp!]
\centering \small
\caption{Annotator recruitment criteria and demographics. Abs. number of mentions across benchmark papers.}
\begin{tabular}[t]{ll}
\toprule
Criterion & \#  \\
\midrule
none        & 15          \\
availability         & 3           \\
task performance     & 6           \\
domain expertise     & 4           \\
other                & 3          \\ \bottomrule
\end{tabular}
\begin{tabular}[t]{ll}
\toprule
Demographic & \#  \\
\midrule
none        & 17          \\
country of origin         & 1           \\
recruitment country     & 3           \\
education     & 3           \\
area of expertise & 5 \\
age & 0 \\
gender & 0 \\
ethnicity & 0 \\
other                & 2          \\ \bottomrule
\end{tabular}
\label{tab:identity_demographic}
\end{table}

\subsubsection{Annotator Demographics}
Finally, we looked for potentially reported demographic details to learn more about the identity or situatedness of those involved in the benchmark creation (relevant for RQ1, as well as RQ3). Out of the 29 benchmark papers involving human annotators, 17 failed to report any demographic information (see Table~\ref{tab:identity_demographic}). Country of recruitment or origin was mentioned for SQuAD, DROP, OpenBookQA, and MATH, exclusively referring to the USA, Canada, or North America in general~\citep{squad, drop, openbookqa, math}. Level of education was mentioned in OpenBookQA (Master's; \citealp{openbookqa}), GPQA (PhD or higher; \citealp{gpqa}), and MMMU (college students; \citealp{mmmu}), which are based on textbook problems or exam knowledge. Information on age, gender, and ethnicity were not identified in the benchmark papers (by internal nor external annotator).
Another indicator for demographic aspects are the author affiliations. We found that those are centered around renown North-American research institutes, universities, and technology firms. 
In sum, 13 of the benchmark papers were co-authored by researchers affiliated to the Allen Institute for Artificial Intelligence (Allen AI) and 8 by researchers affiliated to the University of Washington (UW). 

\subsubsection{Benchmark Motivation}
With reference to RQ2, we were interested to learn what motivated creators to develop their specific benchmarks, and whether or not any of the benchmarks was motivated by an aim to achieve good social representativeness. This was not found to be the case for any of the benchmark papers. Note that the external annotator found SIQA to be aiming for social representativeness since it is framed as a social intelligence benchmark~\citep{siqa}. However, we did not find any evidence that the intention was to improve representativeness in a demographic sense. 

Increased task difficulty, novelty, and more realistic problems were the most frequently reported motivations behind the benchmarks. Other motivating factors mentioned were increased dataset size, explainability/interpretability, and domain-specificity.

\subsubsection{Benchmark Bias and Toxicity}
\label{bias_paper}
In reference to RQ2, we also asked annotators to answer the following question and include evidence for their answer: "Are analyses of aspects related to social bias, representativeness or toxicity in the benchmark dataset reported and, if so, what type of analyses?" The external annotators identified 4 benchmarks as informative in this regard. However, we noticed that they appeared to work on a different understanding of bias than us. For instance, OK-VQA utilizes (non-specific) label balancing to avoid heuristic prediction behavior\footnote{For example, the question  "What season is it?" was mostly accompanied by the answer "Winter" incentivizing the model to default to this answer~\citep{okvqa}.} and for NaturalQuestions, an in-depth analyses of annotation variability was conducted. This indeed can be done in a social bias-sensitive manner~\citep{haliburton2024uncovering}, but in this case the focus was on general annotation quality~\citep{nq}. We count these as uninformative of social bias or toxicity aspects.

We finally identified  3 out of 30 benchmark papers that clearly flag social biases in their data.\footnote{None of the benchmark papers mentioned any toxicity-related metric (full agreement between internal and external annotations).}
The WinoGender bias metric~\citep{rudinger-etal-2018-gender} was applied to models trained on the WinoGrande train split~\citep{winogrande} to verify its relative gender-fairness. 
The QuAC datasheet mentions potential biases towards famous men in its dataset as well as other not further specified biases.\footnote{\url{quac.ai/datasheet.pdf}} 
The GPQA benchmark paper explicitly states that bias was \textit{not} avoided during the dataset creation. The authors "make no claim that GPQA is a representative sample of any population of questions that are likely to come up in the course of scientific practice,"~\cite[][p. 12]{gpqa} and indicate that the crowdworkers tended to default to masculine pronouns when referring to scientists.

An additional keyword matching for the terms "diverse" and "diversity" yielded matches in two thirds of the benchmark papers: Several pay attention to domain or topic diversity~\citep[e.g., ][]{strategyqa, scienceqa, truthfulqa}, 
question or answer diversity~\citep[e.g.,][]{hellaswag, piqa, xquad}, 
as well as lexical diversity~\citep[e.g.,][]{coqa, drop, gsm8k}. 
Yet, again, none of them account for demographic diversity.

\subsubsection{Benchmark Language}
Finally, the language of the benchmark is another factor that can be indicative of a socially relevant form of bias, namely \textit{linguistic bias} (RQ2). All but one of the selected benchmarks were in English, only. Yet, only 12 of the benchmark papers explicitly state this information. In all other cases, we had to derive this information from data examples. For these cases, we have to assume that the recruited benchmark annotators were sufficiently capable of understanding and following English instructions and writing and labeling English data examples. An exception to English as a default, is XQuAD, a multilingual benchmark based on translations of the English SQuAD v1.1 (Spanish, German, Greek, Russian, Turkish, Arabic, Vietnamese, Thai, Chinese, and Hindi; \citealp{xquad}). Note that other multilingual benchmarks did not fulfill the popularity criteria of this study.

\subsection{Benchmark Data Analysis Results}
\label{data_analysis}

 To find more evidence towards answering RQ2 and RQ3, we analyzed distributions of gender, occupation, religion and location properties found for entities across 20 benchmark datasets (see Table \ref{tab:checklist}, Appendix~\ref{paper_checklist}), following the procedure described in Section~\ref{data_analysis_method}.\footnote{The selection of demographic markers reflects dimensions that are frequently discussed in the social bias in NLP literature~\citep{sheng-etal-2021-societal}.} The absolute number of entities differs greatly between benchmarks (see Table \ref{tab:entity_counts},  Appendix~\ref{external_annotations}) due to differences in dataset sizes or the nature of the contents, e.g., HotpotQA (>20k entities) is inherently related to Wikipedia and, thus, highly overlaps with Wikidata, but the commonsense benchmark COPA (<100 entities) does mostly not rely on real-world entities in its examples.\footnote{Example: "The man dropped on the floor. \textit{What happened as a result?}"}  
 \footnote{The entity linker was validated on a small subset of data, consisting of 50 randomly selected items from each benchmark. The first author annotated these random samples by manually listing the Wikidata QIDs associated to entities mentioned therein. The Micro F1 score across benchmarks is .73 (precision=.63, recall=.87).} 

\begin{figure}[htp!]
  \centering
  \includegraphics[width=\linewidth]{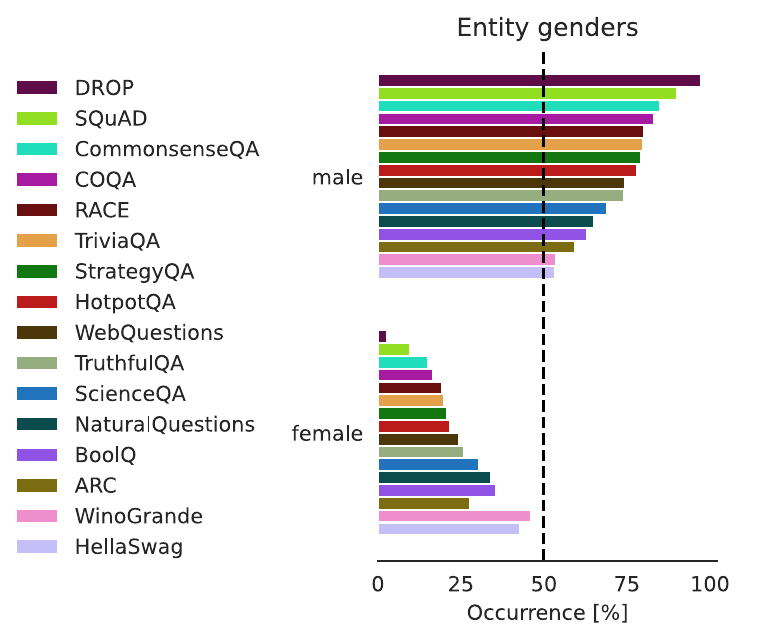}
  \caption{Gender ratio for entities in encyclopedic, commonsense, and scholarly QA \& RC benchmarks.}
  \label{fig:gender}
\end{figure}

\subsubsection{Gender}
Figure~\ref{fig:gender} shows the male-to-female gender ratios across benchmarks. We only included benchmark datasets for which we found more than 30 gender entries. Genders beyond the binary were none or close to none and not illustrated in the plot. The most favorable gender ratios are found in the commonsense benchmarks HellaSwag and WinoGrande (consistent with the low gender bias  reported in the WinoGrande paper;~\citealp{winogrande}). All Wikipedia-based benchmarks, like DROP, SQuAD, or TriviaQA exhibit prominent gender gaps. 
In fact,  DROP  is only based on text passages about male-dominated "National Football League (NFL) game summaries and history articles"~\cite[][p. 2371]{drop}. 

For CommonsenseQA, we only retrieved 28 male and 5 female entities, but we also ran a keyword matching on its question set and found 179 questions containing "he", "man" or "his" and only 49 containing "she", "woman", "her", or "hers". Examples are: "He was working hard on his sculpture, what was he practicing?" and "After she finished washing clothes, what did the woman do with them?" For questions where  gender does not play a role for the task at hand, the dataset creators happened to default more to male subjects. 

Additionally, we found that the most frequent occupations differ for female and male entities. 
For example, for WinoGrande (commonsense), the  top-10 male occupations include several athletic professions, while the  top-10 female occupations are leaning towards entertainment roles (see Figure \ref{fig:occupations}, Appendix~\ref{dataset_biases_vis}).


\subsubsection{Religion}

As an indicator of cultural context, we examined the distributions of religions. 
Firstly, we determined the benchmarks for which 30 or more religion properties were retrieved (ranging between 33 for BoolQ and 652 for TriviaQA). Christianity and instances of Christian religions rank highest across benchmarks. In fact, \textit{Christianity} and/or \textit{Catholicism} are among the top-3 religion labels for 14 out of the 15 benchmarks (see Figure \ref{fig:religions}, Appendix \ref{dataset_biases_vis}). \textit{Islam} is found among the top-3 for HotpotQA, SQuAD, and NaturalQuestions and \textit{Judaism} for BoolQ, WinoGrande, HellaSwag, and TruthfulQA. The other two world religions, \textit{Buddhism} and \textit{Hinduism}, are less represented.  

\subsection{Location}
For the analysis of locations, we again filtered for benchmarks with at least 30 matched location properties. Across encyclopedic, commonsense, and scholarly benchmarks, most coordinates are located around North America and Western Europe. Eastern and Southern regions are less represented. For HotpotQA, TriviaQA, and NaturalQuestions slightly more coordinates are located on the South American, African, and Australian continents compared to the other benchmarks (see Figure \ref{fig:geomap}, Appendix \ref{dataset_biases_vis}). We also retrieved location names associated to entities in the datasets. Again, Western regions are more represented. E.g., for BoolQ and StrategyQA, the most frequently named locations are the \textit{United States} (56\% and 31\%) and the \textit{United Kingdom} (9\% and 15\%), followed by \textit{Canada} (2\%) for BoolQ and \textit{Brazil}, and \textit{Japan} (4\% each) for StrategyQA.

\section{Discussion}
(RQ1) Most of the benchmarks consist of human-authored examples and nearly all involve human annotation. Yet, demographic and recruitment details are rarely reported. 
While the QuAC paper stands out for its comprehensive reporting, several others like MMLU (which is commonly referenced to market flagship models of famous tech firms)\footnote{\url{https://openai.com/index/hello-gpt-4o/}}\footnote{\url{https://www.anthropic.com/news/3-5-models-and-computer-use}} lack all of the details we were looking for.
In a few cases, countries of origin/recruitment are reported (mostly North American). These observations emphasize once more that benchmark creators are not sufficiently sensitized towards the situatedness of their practice~\citep{raji2021ai}. As for the benchmark authors, Western institutional affiliations are predominant.

(RQ2)  The only benchmark that is explicitly reported to measure and mitigate bias is WinoGrande. It utilizes the WinoGender metric to control for (binary) gender bias. Several of the remaining benchmarks datasets are biased regarding gender, occupation, religion, and location of the entities of interest. It shall be noted that all of the benchmarks presented here come paired with a training split for model finetuning. Hence, the biases affect not only evaluation but also training. 
The reliance on Wikipedia (with known representational issues; \citealp{sun2021men, menking2021, tripodi2023mscategorized}) for encyclopedic benchmarks, causes an under-representation of marginalized communities. But also commonsense and scholarly benchmarks were found to default to male and Western examples. 

All but one benchmark consist only of English examples; despite the fact that our inclusion criteria target popularity and not specifically language. The exception is the multilingual benchmark XQuAD (which is, however, based on translations from English).  Less than half of the papers state the dataset language  explicitly, disregarding "the possibility that the techniques may, in fact, be language specific"~\citep[][p. 18]{Bender_2011}. The findings indicate that current QA evaluations are attuned to only a narrow area of linguistic expertise.

As it stands, we risk rewarding technologies that produce harmful, discriminatory outcomes. Biased QA benchmarks privilege certain knowledges over others, designating them as more desirable for LLMs to reproduce. Such LLMs (e.g, as chatbots) widen the gaps in dominant knowledge resources and exacerbate epistemic injustice~\citep{fricker2007epistemic}. 

(RQ3) Previous studies have demonstrated the influence of annotator demographics on annotations~\citep{sap-etal-2022-annotators,pei-jurgens-2023-annotator, al-kuwatly-etal-2020-identifying}. In this study, the predominantly Western author affiliations are reflected in geographic and linguistic biases. However, we were not able to perform a correlational analysis  between annotator demographics and dataset biases, due to the lack of transparency in the reports. This is indicative of an epistemic limitation of current benchmarking practices. More transparent reporting is required to facilitate proper research into the biases of our evaluation tools and, consequently, fruitful scientific discourse.

\paragraph{Recommendations}
Our findings exemplify a "\textit{laissez-faire} attitude"~\cite[][p. 4]{paullada2021data} prevalent in AI dataset creation, which needs to be countered by intentionality and reflexivity. While we acknowledge the growing discourse around better AI evaluation~\citep{wallach2024evaluating,reuel2024betterbench}, we emphasize that the conversation must prioritize social bias alongside validity and transparency.
A first step in conceptualizing a benchmark should be to explicate an ideal distribution and underlying assumptions~\citep{DBLP:conf/acl/ShahSH20, DBLP:conf/acl/BlodgettBDW20}. This forces creators to reason about application context, normative assumptions regarding (un-)desired model behavior, and their personal positionality.  Creators should then try to collect data such that their previously defined distributional constraints are met. This, however, is not easily realized and requires structural changes: there is limited availability of data representing marginalized communities, due to structural societal inequalities~\citep{DBLP:journals/ethicsit/HelmBKG24}. 
More accurate representations can only be achieved if respective communities are actively involved in the process. This must be realized through non-exploitative, \textit{true} participation~\citep{birhane2022power}. We argue that there is not only ethical but also epistemic value in pursuing respective efforts, as this helps to foster \textit{more} representative and generalizable evaluation~\citep{harding1986science}. Limitations and biases are always expected. Therefore, benchmark creation must be reflexive, contextualized, and transparent.

\section{Conclusion}
Our work finds significant limitations regarding transparency and social representativeness in 30 popular QA and RC benchmarks. Many of these benchmarks lack information about annotator demographics, recruitment criteria, and language specificity. Many are linguistically biased and tend to exhibit biases towards entities of certain gender, occupations, religions, and locations. This has objectionable epistemological and ethical implications, e.g., by incentivizing the development of technologies that serve the needs of a privileged few. We highlight the need for rigorous documentation, validation, and representation standards in LLM benchmarking.

\section*{Limitations}
Due to the lack of transparency across benchmarks, we were unable to investigate the causal relationship between the identity of those involved in the benchmark creation and the biases found in the benchmark datasets through statistical testing. 

There is a certain risk that the biases of Wikidata and the entity linker may influence our results. This is hard to avoid in an analysis that utilizes automated processes. Especially for the commonsense and scholarly benchmarks, this is to be considered as a limitation. As for the encyclopedic benchmarks, we assume that this to be less of an issue, because many of them are built on top of Wikidata or Wikipedia (which are content-wise very alike), to begin with. 

 Some time has gone by between conducting this research and the publication of this article. So, it is likely that newer benchmarks would now fall into our selection criteria that we did not consider. Moreover, due to the large annotation efforts required in this study, we had to limit the scope. Therefore, we set strict selection criteria, which happened to exclude multilingual benchmarks. Future work should include a larger number and wider range of benchmarks to allow for more generalizable conclusions. 
 Studies conducted at larger scale should also systematically examine whether benchmarks have become less biased and more transparent over time.

\section*{Acknowledgments}
This work was funded through a Research Fellowship at Weizenbaum Institute, Berlin. It was also supported by the German Research Foundation (DFG) project NFDI4DS under Grant No.: 460234259 and an NVIDIA Academic Hardware Grant.

\bibliography{qa-bench-bias}

\newpage
\appendix

\section{Full Benchmark Paper Checklist}
\label{paper_checklist}

Table \ref{tab:checklist} provides a full checklist regarding reported aspects, category, and inclusion in the dataset analysis across all benchmarks.

\begin{table*}[htp!]
\centering\small
\caption{Checklist of social bias-relevant aspects stated in the benchmark papers \& inclusion in quant. analysis.}
\begin{tabular}{llcccc|c}
\toprule
Year & Benchmark  & Language &  \begin{tabular}[t]{@{}l@{}}Recruitment\\criteria\end{tabular}  & Demographics & \begin{tabular}[t]{@{}l@{}}Social bias\\or toxicity\end{tabular}  & \begin{tabular}[t]{@{}l@{}}Data\\analysis?\end{tabular} \\\midrule
& \textit{Encyclopedic} & & & & &  \\

2018 & QuAC       &   \checkmark   &   \checkmark                                                           &    \checkmark                                                             & \checkmark & - \\

2019 & DROP       &    \checkmark      &    \checkmark                                                            &    \checkmark                                                             &  - & \checkmark\\

2020 & XQuAD      &      \checkmark        &   \checkmark                                                             &  \checkmark                                                               & -  &-\\

2016 & SQuAD      &      -        &    \checkmark                                                            &  \checkmark                                                               & - & \checkmark \\
2018 & HotpotQA & \checkmark &- & -& -& \checkmark\\
2021 & StrategyQA & \checkmark &-&-& -& \checkmark\\
2019 & COQA       &      -        &      \checkmark                                                          &                                                       -          & -& \checkmark \\
2019 & NaturalQuestions&-&-&-&-&\checkmark\\
2017 & TriviaQA &-&-&-&-&\checkmark\\
2019 & BoolQ &-&-&-&-&\checkmark\\
2023 & WebQuestions &-&-&-&-&\checkmark\\\midrule

&\textit{Commonsense} &&&&&\\
2012 & COPA & \checkmark & \checkmark &-&- & \checkmark \\
2021 & WinoGrande &              &     \checkmark                                                           &                                                -                 &  \checkmark&\checkmark\\

2020 & PIQA & \checkmark &-&-&-&- \\

2019 & CommonsenseQA & \checkmark &-&-&- &\checkmark\\
2022 & TruthfulQA &        -      &      \checkmark                                                          &                                                       -          &- &\checkmark \\
2019 & HellaSwag  &       -       &   \checkmark                                                             &                                                        -         & -&\checkmark \\

2019 & SIQA&-&-&-&-& - \\\midrule
& \textit{Scholarly} & &&&&\\
2023 & BioASQ-QA  &  \checkmark     &   \checkmark                                                 &     \checkmark                                                            & - &-\\
2023 & GPQA       &   \checkmark           &    \checkmark                                                            &   \checkmark                                                              & \checkmark &\checkmark \\
2017 & RACE       &     \checkmark         &          -                                                      &    \checkmark                                                             &- &\checkmark  \\
2018 & OpenBookQA &         -     &     \checkmark                                                           &    \checkmark                                                           & - &\checkmark\\

2021 & MATH       &          -    &    \checkmark                                                            &  \checkmark    &  -                                                          & - \\
2022 & ScienceQA&-&-&-&-&\checkmark\\
2021 & MMLU &-&-&-&-&\checkmark\\
2021 & GSM8K &-&-&-&-&-\\
2018 & ARC&-&-&-&-&\checkmark\\\midrule
& \textit{Multimodal} &&&&& \\
2024 & MMMU       &  \checkmark            &  \checkmark                                                              &  \checkmark                                                               &- & -\\
2019 & TextVQA&-&-&-&-&-\\
2019 & OK-VQA&-&-&-&-&-\\
\bottomrule
\end{tabular}

\label{tab:checklist}
\end{table*}

\empty

\section{Benchmark Paper Analysis Ext'd}
\label{external_annotations}
\begin{figure}[htp!]
  \centering
  \includegraphics[width=\linewidth]{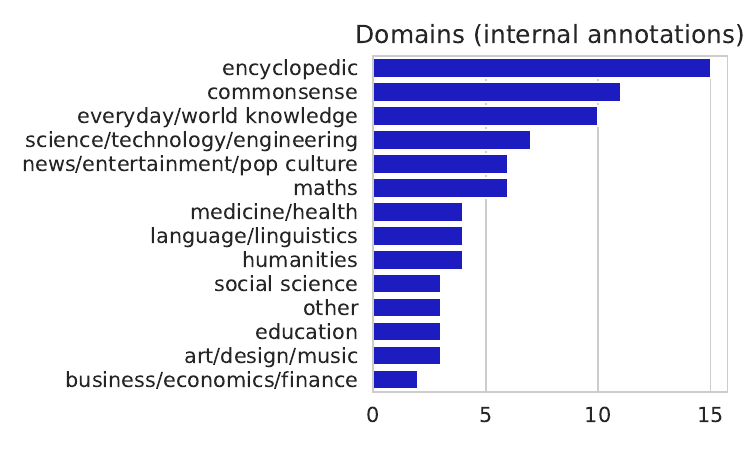}
  \captionof{figure}{Distribution of domains across benchmarks. }
  \label{fig:knowledge_domains}
\end{figure}

Figure \ref{fig:knowledge_domains} provides an overview of the domain/ topic distribution across all benchmarks. Table \ref{tab:motivation_int} lists reported motivations across benchmarks and Table \ref{tab:source_how}  the data sources. Table \ref{tab:identity_demographic_ext} shows the external annotations of annotator recruitment criteria and demographics (internal: Table \ref{tab:identity_demographic}).

\begin{table}[htp!]
\centering \small
\caption{Reported motivations. Abs. counts across papers. Internal (Int.) vs. external (Ext.) annotation.}
\begin{tabular}[t]{lll}
\toprule
Motivation & Int. & Ext.  \\
\midrule
increased difficulty        & 16     &   17  \\
decreased difficulty         & 0     &   1  \\
defining a new task & 10 & 10 \\
more realistic questions    & 9     &    10 \\
better social representativeness     & 0     &   1   \\
other                & 9    &   6   \\ \bottomrule
\end{tabular}
\label{tab:motivation_int}
\end{table}

\begin{table}[htp!]
\centering \small
\caption{Reported data sources. Abs. counts across papers. Internal (Int.) vs. external (Ext.) annotation.}
\begin{tabular}[t]{lll}
\toprule
Source & Int. & Ext.  \\
\midrule
human-authored        & 20     &   20  \\
open access/ web data        & 13     &   14  \\
reusing existing AI/NLP dataset & 8 & 9 \\
exams or textbooks    & 5     &    6 \\
synthetic     & 1     &   1   \\
proprietary/ internal source & 0 & 0 \\
other                & 1    &   2   \\ \bottomrule
\end{tabular}
\label{tab:source_how}
\end{table}

\begin{table}[htp!]
\centering \small
\caption{External annotations of annotator recruitment criteria and demographics. Abs. number of mentions.}
\begin{tabular}[t]{ll}
\toprule
Criterion & \#  \\
\midrule
none        & 14          \\
availability         & 1           \\
task performance     & 7           \\
domain expertise     & 5           \\
other                & 3          \\ \bottomrule
\end{tabular}
\begin{tabular}[t]{ll}
\toprule
Demographic & \#  \\
\midrule
none        & 17          \\
country of origin         & 1           \\
recruitment country     & 2           \\
education     & 4           \\
area of expertise & 3 \\
age & 1 \\
gender & 0 \\
ethnicity & 0 \\
other                & 4          \\ \bottomrule
\end{tabular}
\label{tab:identity_demographic_ext}
\end{table}

\empty

\section{Benchmark Dataset Analysis Ext'd}
\label{dataset_biases_vis}

\begin{figure}[htp!]
  \centering
  \includegraphics[width=\linewidth]{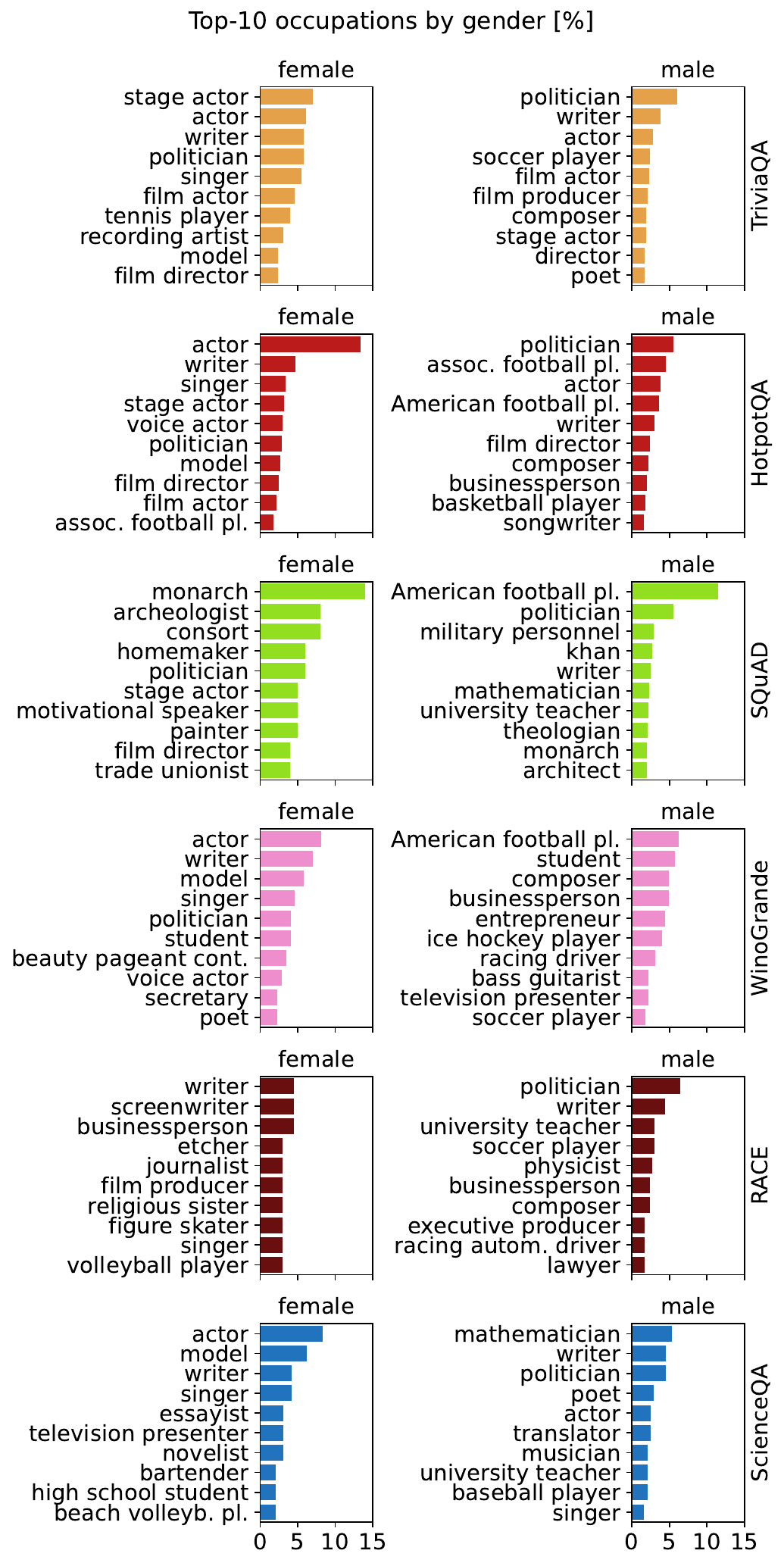}
  \caption{Top-10 occupations by gender across benchmarks (if 300 or more occupations identified).}
  \label{fig:occupations}
\end{figure}

\begin{figure}[htp!]
  \centering
  \includegraphics[width=\linewidth]{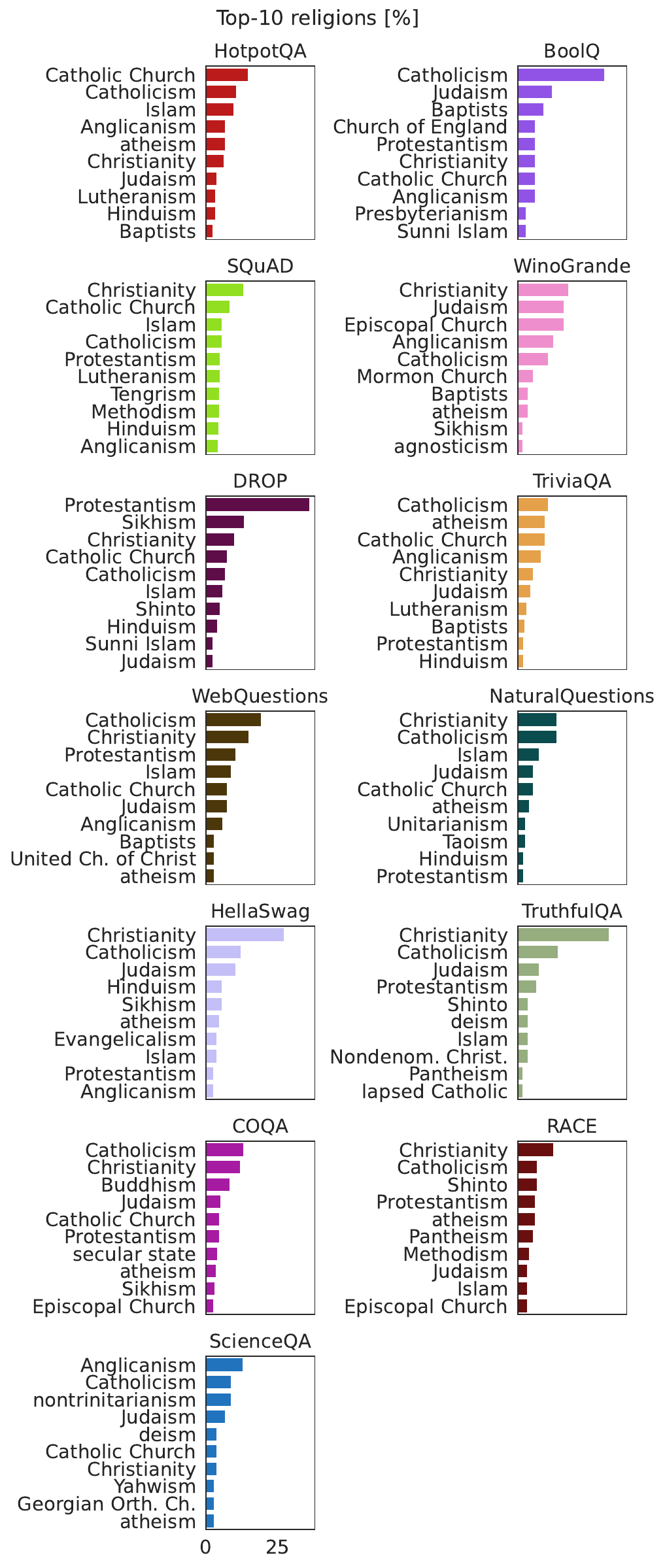}
  \captionof{figure}{Top-10 religions found for entities across benchmarks (if 30 or more instances identified).}
  \label{fig:religions}
\end{figure}

\begin{figure}[htp!]
\centering
\includegraphics[width=\linewidth]{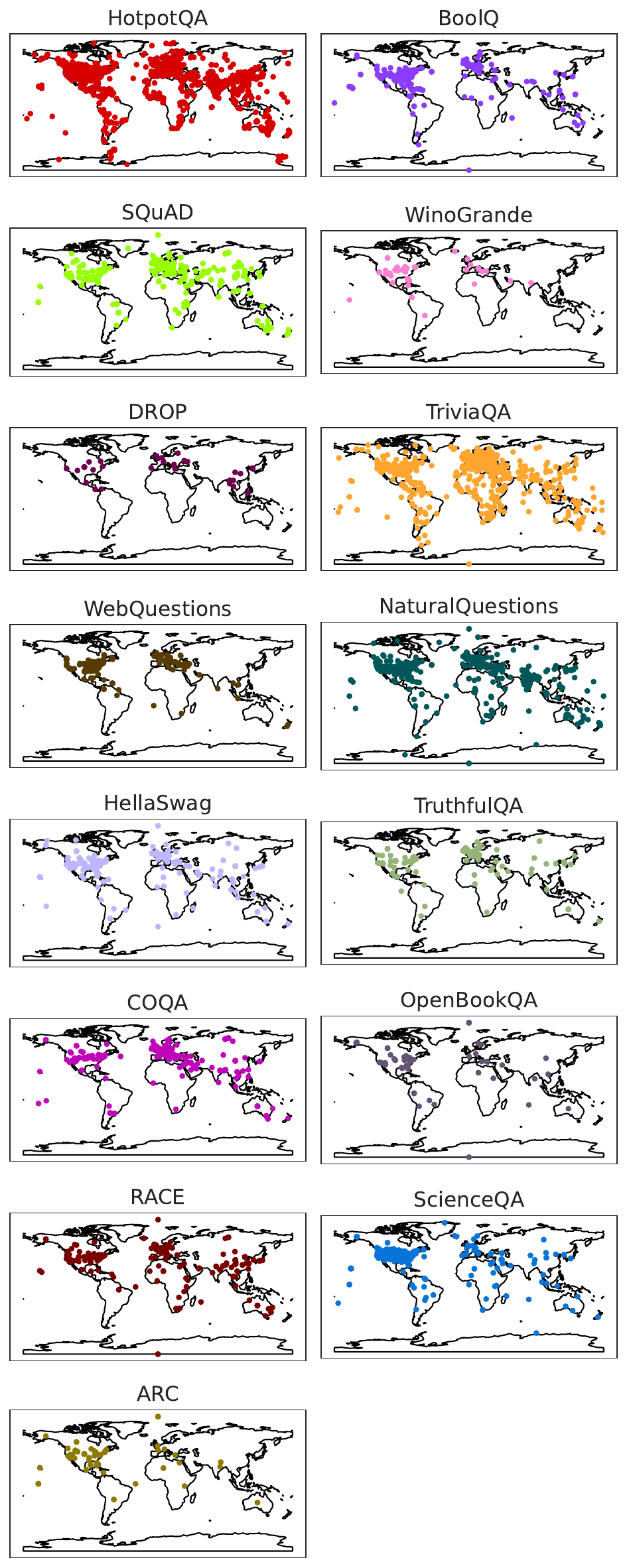}
  \captionof{figure}{Distribution of coordinates found for entities across benchmarks (if 30 or more instances identified).}
  \label{fig:geomap}
\end{figure}

Table~\ref{tab:entity_counts} lists detailed counts of entities extracted using the procedure described in Section \ref{data_analysis_method}. Figures \ref{fig:occupations} and \ref{fig:religions} present relative frequencies of occupations by gender and religion\footnote{Note that we replaced the term "The Church of Jesus Christ of Latter-day Saints" with "Mormon Church" for better proportions of the graph visualization.}  across benchmarks.
Figure~\ref{fig:geomap} illustrates the distributions of coordinates.

\begin{table*}[htp!]
\centering\small
\caption{Detailed list of the numbers of Wikidata entities and associated properties extracted for each benchmark. Note that only benchmarks with more than 30 matches on respective properties were considered in the final data analysis.}
\begin{tabular}{l|c|ccccccc|c}
\toprule
 & \#Entities& && \multicolumn{3}{c}{\#Extracted properties}  & & \\ \midrule
 &               & \begin{tabular}[t]{@{}l@{}}Instance\\of\end{tabular} & Gender & \begin{tabular}[t]{@{}l@{}}Occu-\\pation\end{tabular} & \begin{tabular}[t]{@{}l@{}}Ethni-\\city\end{tabular} & \begin{tabular}[t]{@{}l@{}}Reli-\\gion\end{tabular} & \begin{tabular}[t]{@{}l@{}}Coordi-\\nates\end{tabular} & \begin{tabular}[t]{@{}l@{}}Location\\names\end{tabular} & \begin{tabular}[t]{@{}l@{}}Entity\\linking?\end{tabular}\\\midrule
\textit{Encyclopedic} &&&&&&&&& \\
DROP       &       880     &      804       &    76    &      52    &      14     &     119     &        42     &    411           & - \\
SQuAD      &   10570    &       9462      &   1173     &    1150        &      287     &    610      &     1242        &        4860     &  - \\
HotpotQA &    22189   &          21077   &    6027    &   5684         &       103    &       541   &     3121        &        21103      & - \\
StrategyQA       &  229     &      223       &     48   &       44     &     4      &    18      &      30       &          183     & - \\
COQA      &   1349    &    1194         &     334   &       289     &     136      &     191     &        349     &         1264      &  \checkmark \\
NaturalQu.       &   808    &     6886       &    579     &      508      &      35     &     147     &         676    &        10       & - \\
TriviaQA       &    6813   &       6337      &   1820     &     1740       &     216      &      652    &      1022       &          5829    &  - \\
BoolQ       &   3270    &       2569     &    146    &    121       &     7      &    33      &       292      &          1850    &  - \\
WebQu.  &   755    &     740        &  82      &      75      &       42    &    67      &        213     &          701     & - \\ \midrule
 \textit{Commonsense}   &&&&&&&&& \\
 COPA       &    75   &     50        &   0     &     0       &       0    &      0    &       0      &       5       &  \checkmark \\
WinoGrande       &   799    &     774        &    477    &       356     &     26      &      63    &    56        &      809        & \checkmark \\
Comm.QA  &     208  &     153        &    33    &       25     &     5      &     8     &      26       &      100        & \checkmark \\
TruthfulQA       &     644  &       604      &    62    &       59     &      141     &     107     &      289       &        726      & \checkmark \\
HellaSwag &    3618   &       3228      &    309    &      270      &      201     &     106     &     417        &      2351        & \checkmark \\\midrule
  \textit{Scholarly} &&&&&&&&& \\   
GPQA            &    310   &     274        &    16    &      17      &    13       &     3     &       28      &     99   & \checkmark \\
RACE &  1350   &  1215 &  411 &  370 & 147  &  145 & 349  & 1424  &  \checkmark  \\
OpenB.QA &   282  & 230 &  2 &  2 & 8  & 2  & 57  & 101  &  \checkmark  \\
ScienceQA &  2339   & 1820  & 453  & 346  & 56  &  101 & 554  &  1573 &  \checkmark \\
MMLU &  81   &  69 & 0  & 0  &  0 &  0 &  4 &  6 &  \checkmark \\
ARC &   695  & 570  &  54 &  44 & 18  &  12 &  111 &  338 &   \checkmark    \\
                        \bottomrule
\end{tabular}
\label{tab:entity_counts}
\end{table*}

\end{document}